\documentclass[letterpaper, 10 pt, conference]{ieeeconf}
\IEEEoverridecommandlockouts    
\usepackage{xcolor}

\usepackage{graphics}           
\usepackage{times}              
\usepackage{amsmath}            
\usepackage{amssymb}            
\usepackage{graphicx}
\usepackage{algorithm}
\usepackage[noend]{algpseudocode}
\usepackage{booktabs}
\usepackage{color}
\definecolor{instructioncolor}{rgb}{.5,.5,.5}

\usepackage[font=small]{caption}

\def\figref#1{Fig.~\ref{#1}}
\def\tabref#1{Tab.~\ref{#1}}
\def\eqref#1{Eq.~(\ref{#1})}

\makeatletter
\usepackage{xspace}
\DeclareRobustCommand\onedot{\futurelet\@let@token\@onedot}
\def\@onedot{\ifx\@let@token.\else.\null\fi\xspace}

\def\etal{{et al}\onedot}
\makeatother

\def\etalcite#1{\etal~\cite{#1}}

\usepackage{array}
\newcolumntype{L}[1]{>{\raggedright\let\newline\\\arraybackslash\hspace{0pt}}m{#1}}
\newcolumntype{C}[1]{>{\centering\let\newline\\\arraybackslash\hspace{0pt}}m{#1}}
\newcolumntype{R}[1]{>{\raggedleft\let\newline\\\arraybackslash\hspace{0pt}}m{#1}}

\renewcommand{\b}[1]{\mbox{\boldmath$#1$}}

\renewcommand{\v}[1]{{\b #1}} 

\title{\LARGE \bf Hierarchical Approach for Joint Semantic, Plant Instance, \\ and Leaf Instance Segmentation in the Agricultural Domain}

\author{Gianmarco Roggiolani$^*$ \quad Matteo Sodano$^*$ \quad Tiziano Guadagnino \\[1mm] Federico Magistri  \quad  Jens Behley \quad Cyrill Stachniss
  \thanks{$^*$ Equal contribution.}
  \thanks{All authors are with the University of Bonn, Germany. Cyrill Stachniss is additionally with the Department of Engineering Science at the University of Oxford, UK, and with the Lamarr Institute for Machine Learning and Artificial Intelligence, Germany, T.~Guadagnino was additionally with La Sapienza University of Rome, Italy.}%
  \thanks{This work has partially been funded by the Deutsche Forschungsgemeinschaft (DFG, German Research Foundation) under Germany's Excellence Strategy, EXC-2070 -- 390732324 -- PhenoRob, and by the Deutsche Forschungsgemeinschaft (DFG, German Research Foundation) under STA~1051/5-1 within the FOR 5351~(AID4Crops). 
  }%
}

\begin{document}
\maketitle
\thispagestyle{empty}
\pagestyle{empty}

\begin{abstract}
  %
Plant phenotyping is a central task in agriculture, as it describes plants’ growth stage, development, and other relevant quantities. Robots can help automate this process by accurately estimating plant traits such as the number of leaves, leaf area, and the plant size.
In this paper, we address the problem of joint semantic, plant instance, and leaf instance segmentation of crop fields from RGB data. 
We propose a single convolutional neural network that addresses the three tasks simultaneously, exploiting their underlying hierarchical structure. We introduce task-specific skip connections, which our experimental evaluation proves to be more beneficial than the usual schemes. We also propose a novel automatic post-processing, which explicitly addresses the problem of spatially close instances, common in the agricultural domain because of overlapping leaves. Our architecture simultaneously tackles these problems jointly in the agricultural context. Previous works either focus on plant or leaf segmentation, or do not optimise for semantic segmentation. 
Results show that our system has superior performance compared to state-of-the-art approaches, while having a reduced number of parameters and is operating at camera frame rate.
\end{abstract}

\section{Introduction}
\label{sec:intro}


Sustainable crop farming is fundamental to fulfilling the demand for food, fuel, and fiber while reducing the environmental impact. Plant phenotyping aims to accurately identify plants' growth stages and appearance often to optimize management in the fields or support plant breeders with variety-specific information~\cite{teasdale1998jpa}. The first step is the perception of crops and weeds, which can be automated by robots. Additionally, information about the plant's growth and phenotypic traits can be exploited in automated intervention 
procedures and decision making. One of the popular phenotypic traits is the 
number of leaves each plant has, which is one key aspect of assessing the growth stage and the need of fertilization~\cite{lancashire1991aab}.

In this paper, we propose a solution to simultaneous semantic, plant instance, and leaf instance segmentation of crops.
Given an RGB image recorded by UAVs, our target is to segment crops, individual plants, and their leaves. In the agricultural domain, vision-based approaches mostly target crop-weed classification~\cite{mccool2017ral,milioto2018icra}, plant~\cite{champ2020aps} or leaf segmentation~\cite{weyler2022wacv,weyler2022ral} individually. Additionally, state-of-the-art approaches do not exploit the underlying hierarchical relationship between them.




\begin{figure}[t]
  \centering
  \includegraphics[width=\linewidth]{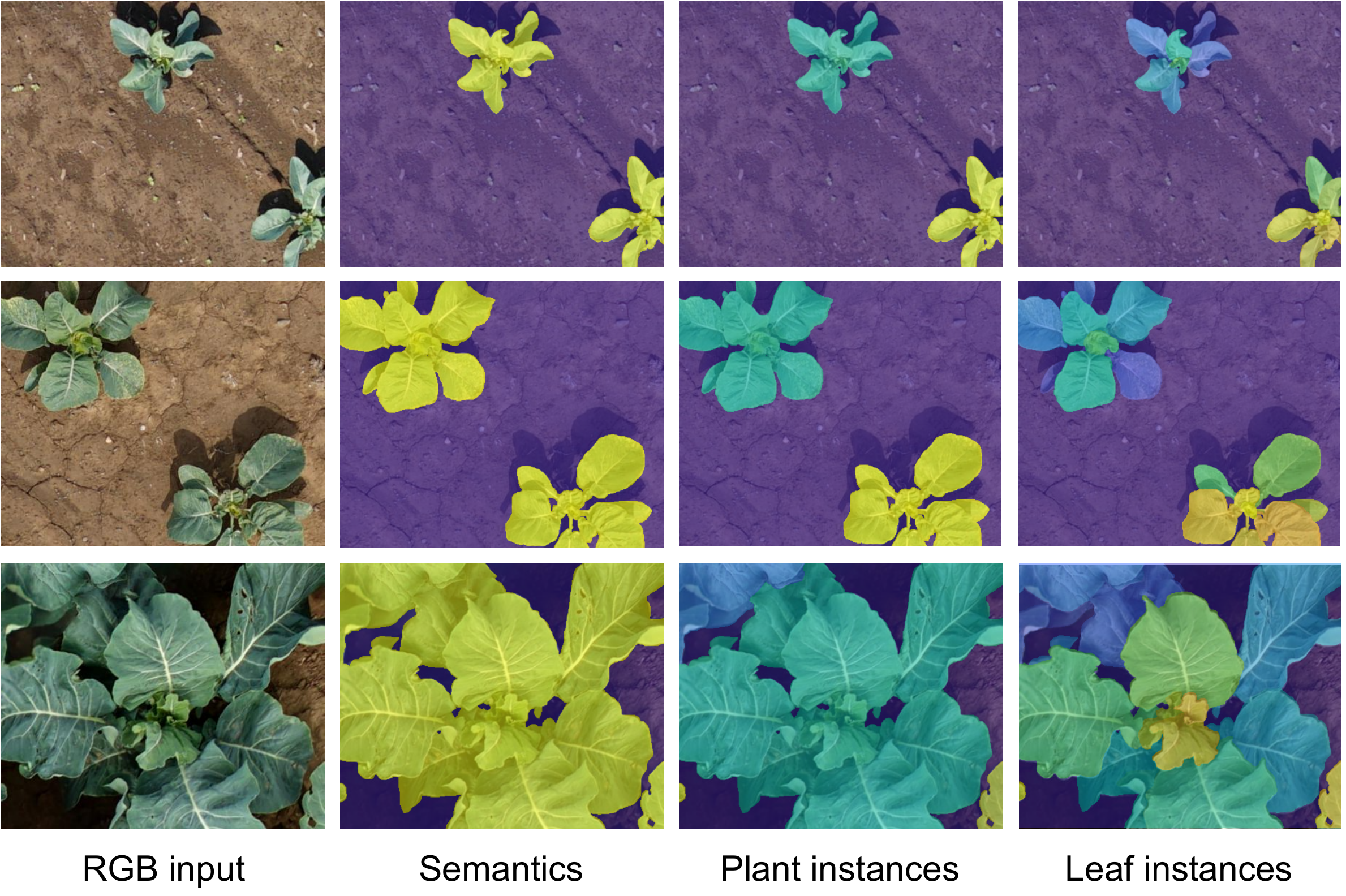}
  \caption{Our approach takes as input RGB images from real fields and provides semantic, plant instance, and leaf instance segmentation. Different instances are shown with different colors.}
  \label{fig:motivation}
  \vspace{-0.25in}
\end{figure}


The main contribution of this paper is a new approach for semantic, plant instance, and leaf instance segmentation that relies on convolutional neural networks (CNN) using RGB data. 
For each pixel in the input, we predict the semantic class and, if it is a crop to which plant and leaf instance it belongs to.
We solve the three tasks jointly, exploiting the underlying task hierarchy by means of a novel design of skip connections. In particular, semantic segmentation can support plant instance segmentation, which can further help leaf instance segmentation.
We additionally propose an automatic post-processing strategy to aggregate the network's outputs and produce the instance mask. Thanks to the structure of our network and the post-processing, our approach yields a pixel-wise semantic, plant instance, and leaf instance segmentation of the image data at the frame rate of a typical camera.
Our experiments suggest that %
(i) our approach can jointly perform semantic, plant instance, and leaf instance segmentation on real-world data;
(ii) our novel scheme for the skip connections better exploits the hierarchical connections between the tasks;
and (iii) our improved post-processing achieves superior performance with respect to common state-of-the-art methods,
while yielding end-to-end inference in real-time.
To support reproducibility, our code is published at \texttt{https://github.com/PRBonn/HAPT}.

\section{Related Work}
\label{sec:related}

Over the last years, we have seen significant progress in the application of vision-based methods for semantic and instance segmentation in real agricultural settings. 
 
Deep learning architectures in the agricultural domain usually target only one specific task, while we address jointly semantic, plant instance, and leaf instance segmentation. This one-shot approach is more efficient, does not require individual networks for each task to run in parallel on the robot, and provides consistent results for the three tasks. 

\textbf{Semantic Segmentation}: In the agricultural domain, many approaches use CNNs to provide a pixel-wise classification of the input image. In Lottes~\etalcite{lottes2018ral}, the authors use as input the sequential images recorded by agricultural robots to exploit the spatial arrangement of the fields. The method by McCool~\etalcite{mccool2017icra} leverages models that can achieve high accuracy and are lightweight to run easily on robotic platforms. Low memory consumption is crucial for real-world applications, together with fast inference time, as also addressed by Milioto~\etalcite{milioto2018icra}. They add the near-infrared (NIR) images as input next to the RGB images and compute multiple vegetation indices as preprocessing to support the training. NIR images are exploited by Bosilj~\etalcite{bosilj2020jfr} as well, but their work focuses on how well the semantic segmentation performance transfers between different crop datasets, to reduce the amount of time and labels needed to train a network on new species. In contrast, Jeon~\etalcite{jeon2018flis} use two architectures in parallel to learn different features that are exchanged during training. Its final result is an ensemble of the outputs. We do not focus on semantic segmentation only, and we experimentally show that performance benefits from sharing information between tasks.

\textbf{Instance Segmentation}: In the agricultural domain, image-based instance segmentation methods aim to detect and segment individual plants or leaves. 
Milioto~\etalcite{milioto2017isprsannals} propose a two-stage approach that first detects single plants and then feeds each one to a CNN classifier to distinguish whether it is a crop or weed. Joint approaches for stem detection~\cite{lottes2020jfr} exist. Several proposals aim to distinguish individual leaves. Morris~\cite{morris2018crv} exploits the differences in texture between the boundaries and the interior of the leaves to segment them through a pyramid CNN. Romera-Parades~\etalcite{romera2016eccv}, instead, focus on the spatial arrangement of the leaves using convolutional long short-term memory units to count them sequentially.
One well-known approach for instance segmentation is Mask R-CNN~\cite{he2017iccv-mr}. It has a two-step procedure where the first step is object detection, and the second produces pixel-wise masks. Though the network is a general purpose one, Champ~\etalcite{champ2020aps} investigated its performance for agriculture. 
The above-mentioned methods only detect plant or leaf instances, but not both jointly, thus limiting the information that we can extract. Weyler~\etalcite{weyler2022wacv} performs plant and leaf segmentation at the same time; they use a bottom-up approach where each plant can be seen as the union of the leaves. In contrast, our approach jointly segments plants and leaves in a top-down fashion. 

\textbf{Plant Phenotyping}: Most of the methods that extract plant traits rely on data acquired in a laboratory, such as the CVPPP Leaf Segmentation Challenge~\cite{scharr2014eccv} where each image presents only one plant.
In this somewhat simplified setting, the approach by Kulikov~\etalcite{kulikov2020cvpr} detects leaves with a two-stage method which first predicts target embeddings with a CNN, and then clusters them. Another common way to deal with leaf counting in literature is by predicting salient points, as in the work from Itzhaky~\etalcite{itzhaky2018bmvc}. They use a CNN to generate a heatmap of leaf keypoints that is fed to a non-linear regression model to predict the number of leaves per plant. Shi~\etalcite{shi2019be} operate in a similar setting, performing semantic and instance segmentation using multiple images of single plants. They combine the predictions of the different viewpoints to 3D point clouds and refine the segmentation of leaves, stems, and nodes. In contrast, the approach presented by Weyler~\etalcite{weyler2022ral} works under real field conditions, detecting the bounding box of single plants and per-plant leaf keypoints. This method, however, only provides coarse keypoints that are not suitable to determine leaf size and shape. In the follow-up work~\cite{weyler2022wacv}, the authors present a model to predict a pixel-wise plant and leaf instance segmentation, allowing for the extraction of relevant plant and leaf traits. 
The main novelty is the introduction of a covariance matrix, optimized by the network, for each instance center, which is crucial for post-processing and final instance detection. Our approach, in contrast, does not use covariance matrices but relies solely on the predictions of offsets and centers for both plants and leaves. Unlike the model from Weyler~\etalcite{weyler2022wacv}, our approach has a dedicated decoder for semantic segmentation, which actually allows us to predict other classes than crops (such as weeds).

\begin{figure*}[t]
  \centering
  \includegraphics[scale=0.87]{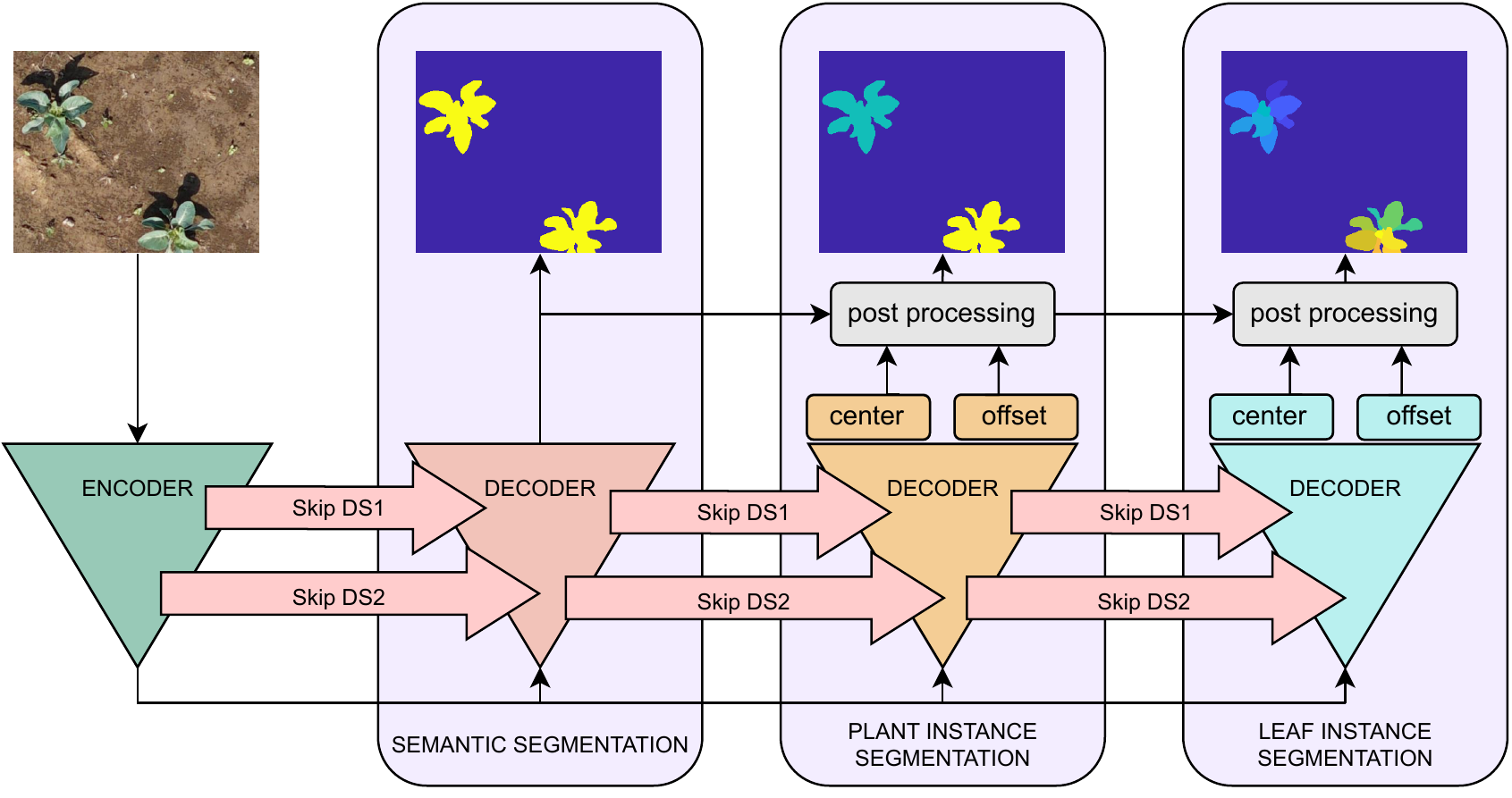}
  \caption{Overview of our architecture. The encoder takes an RGB image as input, process it, and the resulting features are further elaborated by the decoders.
  Skip connections are present in a hierarchical fashion after each downsampling and upsampling block. Skip DS1 and DS2 have size $H/2 \times W/2 \times 64$ and $H/4 \times W/4 \times 128$, respectively. A post processing is necessary to obtain the instance masks.}
  \label{fig:architecture}
  \vspace{-0.2in}
\end{figure*}

\section{Our Approach}
\label{sec:main}

The network that we propose is an encoder-decoder architecture, which takes RGB images \mbox{$I \in \mathbb{R}^{3 \times H \times W}$} as input. The decoders address semantic, plant instance, and leaf instance segmentation, as illustrated in \figref{fig:architecture}. 

\subsection{Architecture}

We use an ERFNet~\cite{romera2018tits} encoder and three ERFNet-based decoders, that allow us to have a lightweight network well-suited for real-time tasks. The semantic segmentation decoder has only one non-bottleneck-1D block after the deconvolutions, while the instance segmentation decoders have two, as defined in the original paper.
Both encoder and decoders use the Gaussian error linear unit (GELU)~\cite{hendrycks2016arxiv} activation function, as suggested in~\cite{liu2022cvpr}.

The \emph{semantic segmentation} decoder has a single output head with depth equal to the number of semantic classes and a softmax activation function. It is trained with the Lovasz-Softmax loss~\cite{berman2018cvpr}, denoted as $\mathcal{L}_{\mathrm{sem}}$. 

The \emph{instance segmentation} decoders have two heads each,
for centers and offsets prediction. The center prediction heads
have an output depth of 1 and a sigmoid activation function
to predict pixel-wise probabilities of being a center. We define the center of an object as the internal pixel closest to its median point. The center predicition is optimized with a binary focal loss $\mathcal{L}^i_{\mathrm{cen}}, \ i \in \{p, \, l\}$~\cite{milioto2019icra-fiass}, where $p$ and $l$ stand for plants and leaves.
The offset prediction heads have an output depth of 2 since they predict offset images in both the $x$ and $y$ directions, and are optimized with L1 losses $\mathcal{L}^i_{\mathrm{off}}, \ i \in \{p, \, l\}$. 

Thus, the final loss function $\mathcal{L}$ is given by
\begin{equation}
\label{eq:loss}
  \begin{split}
  \mathcal{L} = & \, w_1 \, \mathcal{L}_{\mathrm{sem}} \, +  \, w_2 \, \mathcal{L}^p_{\mathrm{cen}} \, +  \, w_3 \, \mathcal{L}^l_{\mathrm{cen}} \, + \\ & \, w_4 \, \mathcal{L}^p_{\mathrm{off}}
  \, +  \, w_5 \, \mathcal{L}^l_{\mathrm{off}},
  \end{split}
\end{equation}
where $w_i$ are scalar weights for the different terms.

\subsection{Skip Connections}

Skip connections are fundamental in several architectures to ensure feature reusability and solve the degradation problem of deep models. They skip one or more layers and provide a direct gradient flow from late to early stages. This preserves low-level spatial information usually lost during downsampling. 
The common usage of skip connections in segmentation models is inspired by Ronneberger~\etalcite{ronneberger2015micc}, where the higher-resolution feature maps of the encoder are concatenated with the features maps of the decoder. 

In this work, we suggest a new scheme for skip connections that takes into account the relations between the different tasks we address. We propose to connect directly different decoders, rather than encoder and decoders only, to improve segmentation performance. 
Our skip connections propagate feature maps of dimensions $H/2 \times W/2 \times 64$ and $H/4 \times W/4 \times 128$, as shown in~\figref{fig:architecture}. In particular:
\begin{enumerate}
  \item For the semantic segmentaton task, we keep the skip connections from the encoder to the decoder, since spatial information coming from the features extrated at higher-resolutions helps the decoder to correctly classify each pixel.
  \item Plant instance segmentation aims to distinguish each pixel classified as crop in a specific instance, so the information coming from the semantic task is more helpful than the features coming from the encoder only. Thus, we sum the contribution coming from the semantic segmentation decoder, to focus on the relevant regions of interest.
  \item The objective of leaf instance segmentation is to discriminate individual leaves in each plant. To achieve this, knowing the position of each distinct plant is more helpful than just knowing where crops are or feature maps from earlier stages. Thus, analogously to before, we augment the skip connections with the contribution from the plant instance decoder.
\end{enumerate}
This newly-proposed skip connection scheme directly exploits the underlying hierarchy between the tasks, designed to realize a meaningful transfer of features from one branch (either encoder or decoder) to the other. Extensive experiments reported in Sec.~\ref{sec:expskip} suggest that these task skip connections lead to superior perfomances.


\begin{figure*}[t]
  \centering
  \includegraphics[width=0.95\linewidth]{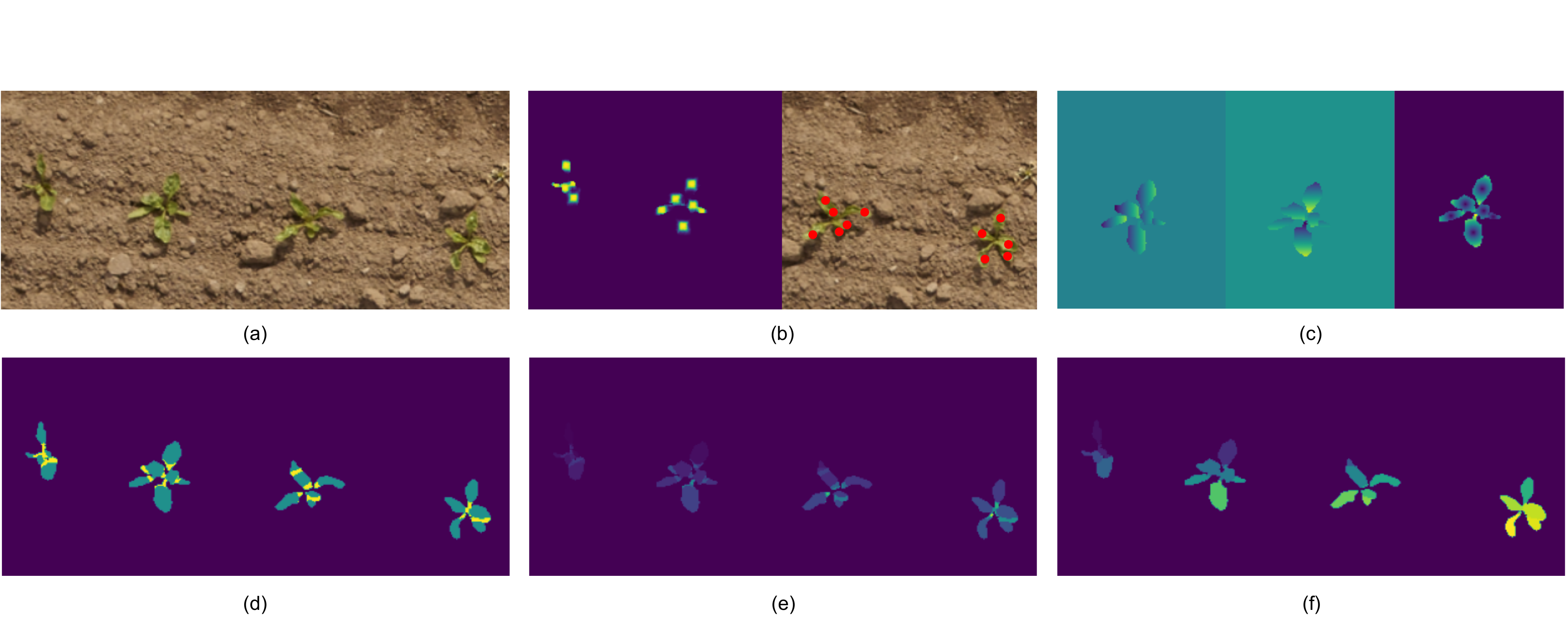}
  \caption{Visual workflow of our 3-step post processing pipeline: (a)~the input image, (b)~the predicted leaves centers before and after non-maximum suppression, (c)~the predicted offsets in the x and y direction, and distance map $\hat{D}$ computed from them for the second plant from the left, (d)~in green the pixels assigned during step 2, in yellow those assigned in step 3, (e)~leaf instance segmentation after step 2, (f)~final leaf instance segmentation after step 3.}
  \label{fig:postprocessing}
  \vspace{-0.2in}
\end{figure*}

\subsection{Post-Processing}

Our automatic post-processing is a 3-step procedure. We show each step in detail for an example image in \figref{fig:postprocessing} (a). The first step is inspired by Panoptic DeepLab~\cite{cheng2020cvpr}, and has the objective to extract a single center for each individual object. Specifically,
we take the center prediction coming from the decoder and filter it with the predicted semantic mask to discard 
any center that does not belong to the class of interest. Since the center prediction head usually outputs blobs around the desired center, we perform a non-maximum suppression operation in order to reduce each blob to a single pixel, as shown in \figref{fig:postprocessing} (b). 

Afterwards, we need to assign each pixel to its center, that defines the individual instance. However, the offsets could point to regions of space close to more than one center. In the second step, we assign only those pixels whose offsets point to a single center. To this end, we build an image of coordinates, where each pixel $\v{p}_{i,j} \in \mathbb{R}^2$ is a vector of values $(i,j), i \in \left[ 1, H \right], j \in \left[1, W \right]$. We compute the Euclidean distance between this image and every center~$c$, producing for each center a distance map $D^c$. Then, we compute a predicted distance map $\hat{D}$ from the offsets, as shown in \figref{fig:postprocessing}~(c). When the offsets point close to a center~$c$, we expect the predicted distance map to be similar to $D^c$. 
Thus, defining a distance threshold $\tau$, one pixel $\v{p}_{i,j}$ is assigned to the instance with center $c$ if
  
\begin{equation}
  \label{eq:pp}
  \left|\left| D^c_{\v{p}_{i,j}} - \hat{D}_{\v{p}_{i,j}} \right|\right| \, \leq \tau,
\end{equation}
holds for that instance only. In \figref{fig:postprocessing} (d), we see the pixels that are now assigned and those that are not. The instance mask at this point is displayed in \figref{fig:postprocessing} (e). 

The third step takes care of the pixels that were not assigned to any instance. This can happen if their offset points too far from every extracted center or close to more than one. In this case, we use a voting mechanism. We compute the instance label that occurs the most between the $N$ closest neighbours and assign it to the current pixel. The outcome of the automatic post processing can be seen in \figref{fig:postprocessing} (f).
To enforce consistency between the masks, we filter all post-processing results with the semantic segmentation masks.



\begin{figure*}[t]
  \centering
  \includegraphics[width=\linewidth]{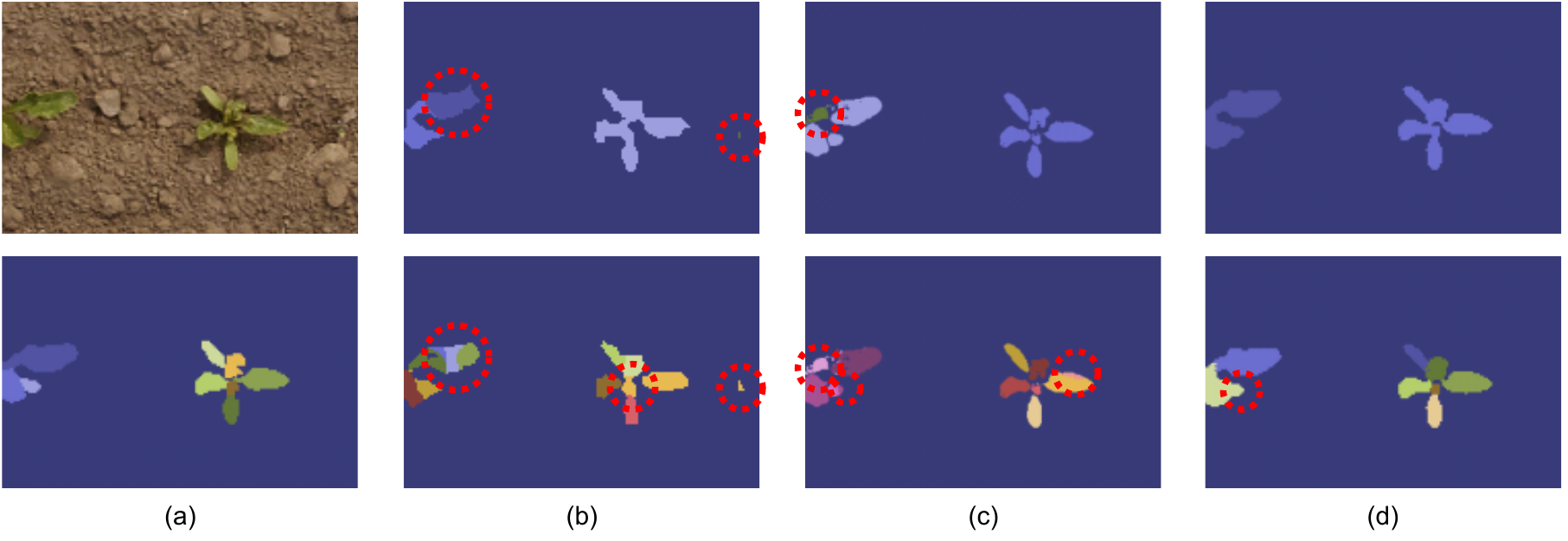}
  \caption{Qualitative results on the SugarBeets dataset for plant (top) and leaf (bottom) instance segmentation. We show the input image and ground truth leaf instances in~(a), and results from PD-S~(b), Weyler~(c), and our approach~(d). Errors in the red circles. }
  \label{fig:bosch}
  \vspace{-0.1in}
\end{figure*}

\section{Experimental Evaluation}
\label{sec:exp}

%

%
We present our experiments to show the capabilities of our method for joint semantic, plant instance, and leaf instance segmentation of RGB data. The results of our experiments support our key claims, which are:
(i) our approach can jointly perform semantic, plant instance, and leaf instance segmentation in one-shot on real-world data,
(ii) our novel scheme for the skip connections better exploits the hierarchical connections between the tasks;
and (iii) our improved post-processing achieves superior performance with respect to common state-of-the-art methods.

\subsection{Experimental Setup}

\textbf{Datasets}. We test our method on two RGB datasets: a sugar beets dataset introduced by Weyler~\etalcite{weyler2022wacv} (denoted as SugarBeets in the following) and
GrowliFlower~\cite{kierdorf2022jfr}. SugarBeets is 
composed of 1,316 images with a resolution of $512 \, \mathrm{px} \times 1024 \, \mathrm{px}$. 
The images are recorded with an UAV equipped with a PhaseOne iXM-100 
camera mounted in nadir view. GrowliFlower is a dataset of cauliflower images. It is composed of 
2,198 images with a resolution of $368 \, \mathrm{px} \times 448 \, \mathrm{px}$. The images are recorded with an UAV equipped with 
a Sony A7 rIII RGB camera and a MicaSense 5CH for multispectral image data. Both datasets provide an official data split that we adopt.

\textbf{Metrics}. For semantic segmentation,
we compute the intersection over union (IoU)~\cite{everingham2010ijcv} of the ``crop'' class. 
For the plant and leaf instance segmentation, we evaluate our method by means of the panoptic
quality (PQ)~\cite{kirillov2019cvpr-ps}.

\textbf{Training details and parameters}. In all experiments, we use AdamW~\cite{loshchilov2017arxiv} without weight decay 
with an initial learning rate of $5\cdot 10^{-4}$ for the encoder and the semantic decoder and $8\cdot10^{-4}$ for the instance decoders, for 500 epochs. 
We initialize our network with the Xavier initialization~\cite{glorot2010aistats}.
The batch size is set to 1. We resize images from the SugarBeets dataset to $256\,\mathrm{px} \times 512\,\mathrm{px}$ to keep the aspect ratio. No resize 
is applied to the GrowliFlower dataset. Additionally, we set $w_1 = 1$, $w_2 = w_3 = 0.1$, and $w_4 = w_5 = 50$ in~\eqref{eq:loss}, while in the post-processing we use number of neighbors $N = 5$, grouping threshold $\tau = 6$ for the plant instance segmentation and 2 for the leaf instance segmentation. We tuned all hyperparameters on the validation sets.






The first experiment evaluates the performance of our approach and its outcomes support the 
claim that we can jointly provide pixel-wise semantic, plant instance, and leaf instance segmentation.  
\tabref{tab:bosch} and \tabref{tab:growliflower} 
show the IoU for the crops, and the panoptic quality for both plant~(PQ$_P$) and leaf~(PQ$_L$) instances. 
We also report the number of parameters of the networks and the end-to-end frame rate of each method at inference time (FPS).
We compare against Mask R-CNN~\cite{he2017iccv-mr} (denoted as MR), which is a common approach in the agricultural domain,
and Panoptic DeepLab~\cite{cheng2020cvpr} (denoted as PD), which is a state-of-the-art model for panoptic segmentation. 
We use three variants of Panoptic Deeplab, with different backbones: a small model that uses MobileNetV2~\cite{sandler2018cvpr}
(called PD-S in the tables), a medium-size model with ResNet50~\cite{he2016cvpr} (PD-M), and a 
big-size model with Xception65~\cite{chollet2017cvpr} (PD-L). 

All these baselines, however, can only address one instance 
segmentation task at a time and, thus, they need to be trained for either plants-only or leaves-only. 
We also compare with the work from Weyler~\etalcite{weyler2022wacv} (denoted as Weyler), which addresses both, plant and leaf instance 
segmentation. 

\begin{figure*}[t]
  \centering
  \includegraphics[width=0.95\linewidth]{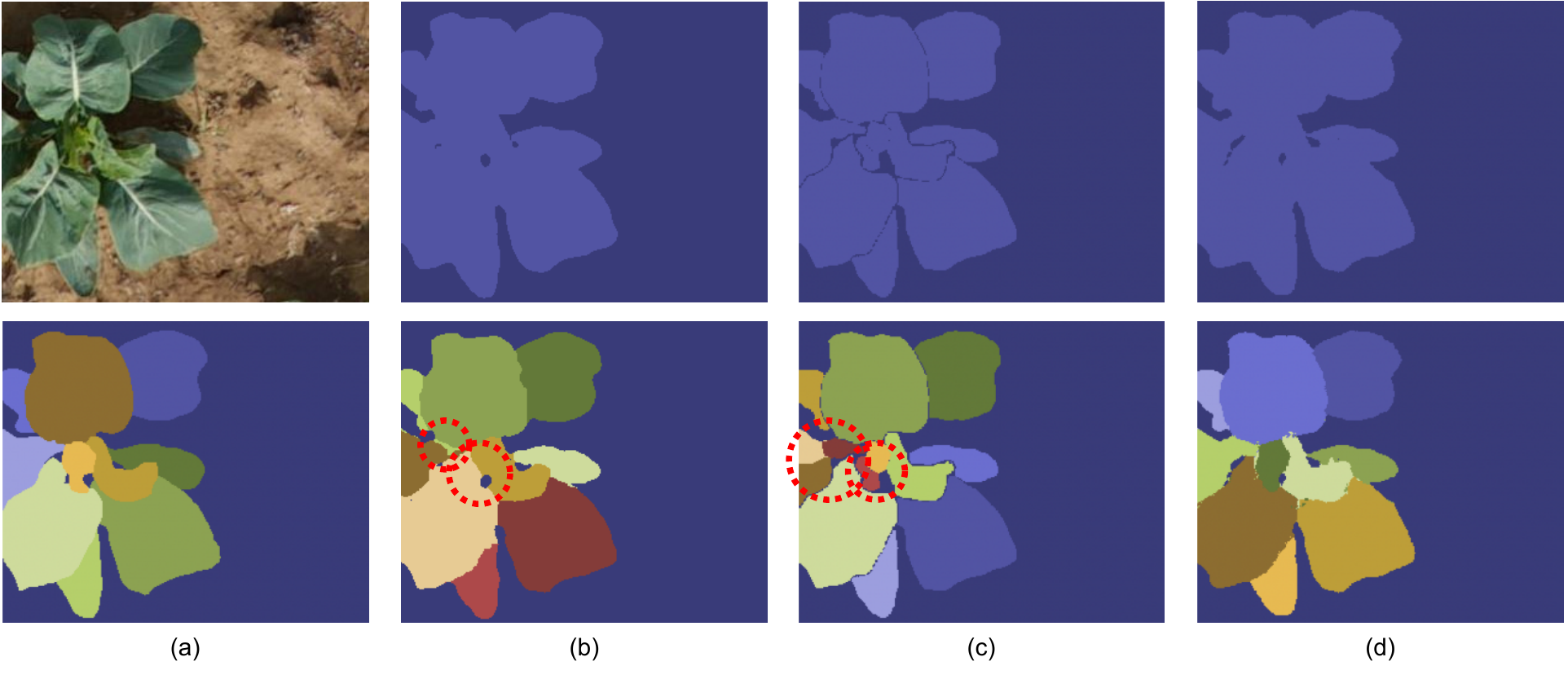}
  \caption{Qualitative results on the GrowliFlower dataset for plant (top) and leaf (bottom) instance segmentation. We show the input image and ground truth leaf instances in~(a), and results from PD-S~(b), Weyler~(c), and our approach~(d). Errors in the red circles. }
  \label{fig:growli}
  \vspace{-0.1in}
\end{figure*}

\setlength\tabcolsep{5.3pt}\begin{table}[t]
  \caption{Performance of baselines and our model on the test set of the SugarBeets dataset. P and L stand for plant and leaf instance segmentation, respectively. Best results in bold.}
  \centering
  
  \begin{tabular}{cccccccc}
    \toprule
    Model                                               & P       & L        & IoU~$\uparrow$          & PQ$_P$~$\uparrow$        & PQ$_L$~${\uparrow}$         & Params        & FPS~${\uparrow}$    \\
    \midrule
    MR~\cite{he2017iccv-mr}         & \checkmark &         & 46.2          & 47.8         & -               & 43.9M         & 13.5 \\    
    PD-S~\cite{cheng2020cvpr}      & \checkmark &    & 75.4          & 69.4         & -               & 7.7M          & 93.5 \\
    PD-M~\cite{cheng2020cvpr}     & \checkmark &     & 75.5          & 69.8         & -               & 55.3M         & 4.7 \\
    PD-L~\cite{cheng2020cvpr}     & \checkmark &     & 76.4          & 71.1         & -               & 69.6M         & 48.4 \\
    \midrule
    MR~\cite{he2017iccv-mr}          & & \checkmark        & 64.9          & -            & 53.6            & 43.9M         & 13.4 \\    
    PD-S~\cite{cheng2020cvpr}     & & \checkmark      & 75.4          & -            & 50.8            & 7.7M          & 93.7 \\
    PD-M~\cite{cheng2020cvpr}     & & \checkmark      & 76.7          & -            & 54.4            & 55.3M         & 49.1 \\
    PD-L~\cite{cheng2020cvpr}    & & \checkmark       & 76.3          & -            & 52.9            & 69.6M         & 48.5 \\
    \midrule
    Weyler~\cite{weyler2022wacv}          & \checkmark & \checkmark & 75.3          & 72.3         & 63.1            & 2.25M         & 0.14 \\
    Ours                            & \checkmark & \checkmark     & \b{79.3}      & \b{76.2}     & \b{63.5}        & 2.4M          & 26.3 \\
    \bottomrule
  \end{tabular}
  \label{tab:bosch}
  \vspace{-0.15in}
\end{table}

\setlength\tabcolsep{5.3pt}\begin{table}[t]
  \caption{Performance of baselines and our model on the test set of the GrowliFlower dataset. P and L stand for plant and leaf instance segmentation, respectively. Best results in bold.}
  \centering

  \setlength\tabcolsep{5.3pt}\begin{tabular}{cccccccc}
    \toprule
    Model                                               & P       & L        & IoU~${\uparrow}$          & PQ$_P$~${\uparrow}$        & PQ$_L$~${\uparrow}$         & Params        & FPS~${\uparrow}$    \\
    \midrule
    MR~\cite{he2017iccv-mr}         & \checkmark &         &  25.4          &  27.9         & -               & 43.9M         & 9.6 \\    
    PD-S~\cite{cheng2020cvpr}  & \checkmark &        & 83.1          & 69.9         & -               & 7.7M          & 43.4 \\
    PD-M~\cite{cheng2020cvpr}  & \checkmark &        & 82.0          & 68.0         & -               & 55.3M         & 47.6 \\
    PD-L~\cite{cheng2020cvpr}   & \checkmark &       & 82.7          & 69.4         & -               & 69.6M         & 23.8 \\
    \midrule
    MR~\cite{he2017iccv-mr}           & & \checkmark         & 53.8          & -            & 41.0            & 43.9M         & 16.2 \\    
    PD-S~\cite{cheng2020cvpr}    & & \checkmark        & \b{84.4}          & -            & 58.8            & 7.7M          & 76.5 \\
    PD-M~\cite{cheng2020cvpr}    & & \checkmark        & 80.2          & -            & 43.4            & 55.3M         & 41.6 \\
    PD-L~\cite{cheng2020cvpr}    & & \checkmark        & 82.8          & -            & 50.1            & 69.6M         & 30.3 \\
    \midrule
    Weyler~\cite{weyler2022wacv}   & \checkmark & \checkmark  & 65.8          & 67.8         & 69.4            & 2.25M         & 0.25 \\
    Ours                               & \checkmark & \checkmark  & 80.2      & \b{89.2}    & \b{71.0}        & 2.4M          & 20.7 \\
    \bottomrule
  \end{tabular}
  \label{tab:growliflower}
  \vspace{-0.15in}
\end{table}
\subsection{Experiments on Double Panoptic Segmentation}
\label{sec:expskip}
Interestingly, the models that tackle all tasks are also the smallest one in terms of number of parameters, 
which makes it more suitable to run on resource-constrained robotic systems. The semantic segmentation decoder, 
which is our first output and filters the following predictions, is the reason behind the extra parameters compared to Weyler.  
Our approach is suitable for real-time operations with a frame rate that exceeds $20$\,Hz. 
All baselines have worse segmentation performance on all the tasks. Additionally, most of them need two models 
to perform all tasks that we tackle with our network, which also means that the same RGB image needs to 
pass through two models that do not share parameters and two post-processing operations that does not ensure consistency 
of the results. The only baseline that addresses both, the plant and leaf instance with one network is
Weyler~\etalcite{weyler2022wacv}, which is not suitable for real-time operations due to its relatively low framerate of $0.14$\,Hz on SugarBeets and $0.25$\,Hz on GrowliFlower.

In sum, our model with specifically-designed skip connections and novel automatic post-processing operations outperforms 
state-of-the-art architectures on all tasks on the SugarBeets dataset. On GrowliFlower, our model substantially outperforms all baselines on instance segmentation tasks. Additionally, our model is able to run at the frame rate of common RGB cameras. 
Qualitative results are shown in~\figref{fig:bosch} for SugarBeets, and in~\figref{fig:growli} for GrowliFlower.



\subsection{Ablation Studies}

In this section, we provide ablations to show the improvements provided by the
skip connections scheme and the post-processing operations. We perform all ablations on the validation set of the SugarBeets dataset. 

We compare our novel skip connection scheme against other ways to connect them and show the results in \tabref{tab:ablation_skips_postproc}. In particular, 
we use the exact same network as the one we propose with no skip connections~(A), typical 
encoder-decoder skip connections~\cite{ronneberger2015micc}~(B), hierarchical skip connections with no gradient flow~(C), skip connections without summing the contribution from the encoder~(D). 
When we do not use any skip connection the panoptic qualities are noticeably lower, 
because the corresponding decoders have no help from previous features. In the case of encoder-decoder 
skip connections, the panoptic qualities are better since the decoders get features from the encoder that has to ``compromise'' between all tasks, harming the semantic segmentation. Interestingly, we notice
no improvement from the hierarchical skip connections with no gradient flow, where skip connections are detached and thus they do not participate to the backward pass, since the feature 
flow does not play any role in the optimization, leading to suboptimal performance. 
On the other hand, hierarchical skip connections without the encoder contribution substantially improve performance with respect to the skip connections from the encoder only. This suggest that decoder features are more relevant than restoring features from the encoder when it comes to tasks that present an underlying hierarchical structure. 
In the same table, we also show the performance of our best model with the post processing from Panoptic DeepLab~(E). Clearly, a different post-processing does not play any role in the semantic segmentation, and our post-processing substantially improves instance segmentation, especially when clustering leaves~(F).

Our last ablation study focuses on the standard panoptic segmentation problem: semantic and (a single) instance segmentation. We keep the plant instance decoder in order to maintain the hierarchy of our approach. As we can see in \tabref{tab:ablation_skips}, we evaluate our skip connections scheme against the commonly-used encoder-decoder strategy coming from UNet~\cite{ronneberger2015micc}~(G). 
The experiment confirms that our skip connection scheme~(H) exploits the hierarchy between the tasks better, leading to superior segmentation performance. 


\setlength\tabcolsep{5.0pt}\begin{table}[t]
  \caption{Comparison between different skip connections and different post processings. PD stands for Panoptic DeepLab. Best results in bold.}
  \centering

  \begin{tabular}{ccccccc}
    \toprule
    & Skip Connections    & Attached   & Post-proc       & IoU           & PQ$_P$        & PQ$_L$        \\
    \midrule
    A & None            &       & Ours                 & 84.4         & 75.5          & 65.1          \\    
    B & Encoder          &  \checkmark   & Ours                 & 83.3         & 79.4          & 65.6          \\    
    C & Task + Encoder  &      & Ours                 & 83.2         & 78.9          & 65.6          \\  
    D & Task       & \checkmark & Ours                 & 84.4     	 & 81.1      	 & 66.0 
\\    
    E & Task + Encoder   &   \checkmark    & PD~\cite{cheng2020cvpr}   & \b{84.5}      & 79.0          & 47.2          \\    
    F & Task + Encoder    &  \checkmark    & Ours                 & \b{84.5}     & \b{81.7}      & \b{67.8}          \\    
    \bottomrule
  \end{tabular}
  \label{tab:ablation_skips_postproc}
\end{table}

\setlength\tabcolsep{5.0pt}\begin{table}[t]
  \caption{Comparison between our skip connections and the UNet~\cite{ronneberger2015micc} encoder-decoder scheme for the standard panoptic segmentation task (semantic + single instance). Best results in bold.}
  \centering

  \begin{tabular}{cC{3cm}cccc}
    \toprule
    & Skip Connections   & Attached    & Post-proc     & IoU           & PQ$_P$     \\
    \midrule
    G & Encoder          & \checkmark      & Ours               & 85.1          & 81.0        \\    
    H & Task + Encoder   & \checkmark       & Ours               & \b{85.8}      & \b{82.6}    \\    
    \bottomrule
  \end{tabular}
  \label{tab:ablation_skips}
  \vspace{-0.1in}
\end{table}

\section{Conclusion}
\label{sec:conclusion}

In this paper, we introduced a novel approach for joint hierarchical semantic, plant instance, 
and leaf instance segmentation of RGB images.
Our method exploits the inner task hierarchy by means of a specifically-designed 
skip connection scheme, while improving instance segmentation results with a novel 
post processing operation.
This allows us to successfully outperform current state-of-the-art approaches while 
having a fast model that is well-suited for operation on a mobile robot. 
We implemented and evaluated our approach on different datasets
and provided comparisons to other existing techniques and supported
all claims made in this paper. The experiments suggest that exploiting task 
hierarchy is crucial for effective segmentation results.



\bibliographystyle{plain_abbrv}

\bibliography{glorified,new}

\end{document}